\documentclass[letterpaper, 10 pt, conference]{ieeeconf}

\IEEEoverridecommandlockouts
\usepackage{url}
\usepackage{graphicx}
\usepackage{amsmath,amssymb,amsfonts}
\usepackage{array}
\usepackage{tabularx}
\usepackage{booktabs}
\usepackage{multirow}
\usepackage{makecell}
\usepackage{xspace}
\usepackage[normalem]{ulem}
\usepackage[table]{xcolor}
\newcolumntype{L}[1]{>{\raggedright\arraybackslash}p{#1}}
\newcolumntype{C}[1]{>{\centering\arraybackslash}p{#1}}
\newcolumntype{Y}{>{\centering\arraybackslash}X}

\newcommand{\best}[1]{\textbf{#1}}
\newcommand{\second}[1]{\underline{#1}}
\newcommand{\NA}{--}

\title{\LARGE \bf
DRAgent: Discriminative Reasoning Agent for Referring Expression Segmentation
}

\author{Yujie Qi$^{1}$ and Luyan Zhang$^{2,*}$%
\thanks{$^{1}$School of Computer Science and Technology, Hangzhou Dianzi University, Hangzhou, 310018, China.}%
\thanks{$^{2}$Khoury College of Computer Sciences, Northeastern University, Vancouver, BC V6B 1Z3, Canada.}%
\thanks{$^{*}$Corresponding author: zhang.luya@northeastern.edu}%
}

\begin{document}

\maketitle
\thispagestyle{empty}
\pagestyle{empty}

\begin{abstract}

Referring Expression Segmentation (RES) aims to generate a pixel-level mask for the object specified by a language expression. Recent methods based on multimodal large language models (MLLMs) often rely on one-pass coordinate prediction for visual localization, which serializes continuous spatial locations as discrete text tokens and may lead to localization bias and alignment errors. To address these issues, we propose DRAgent, an MLLM-driven discriminative reasoning (DR) framework for RES. Instead of requiring the MLLM to generate localization coordinates, DRAgent first constructs a detector-generated candidate space and then uses the MLLM as a visual-semantic target discriminator. Specifically, the MLLM performs reliable target selection among potential distractors through a two-stage DR mechanism, which first screens high-recall candidates and then performs instance-wise verification. The selected target box is subsequently used as a spatial prompt for a foundation segmentation model to produce the final pixel-level mask. Furthermore, we construct a self-consistency-filtered reasoning-chain data pipeline for LoRA-based fine-tuning, providing more reliable supervision for enhancing the MLLM’s discriminative reasoning capability. Experiments demonstrate that DRAgent achieves competitive performance on RefCOCO, RefCOCO+, and RefCOCOg. 

\end{abstract}

\section{Introduction}


Referring Expression Segmentation (RES) aims to generate a pixel-level mask for the object specified by a natural language expression~\cite{LAVT,ReLA}. In autonomous systems such as industrial or assistive robotics, it serves as a critical bridge between perception and control, where precise identification is the prerequisite for subsequent task execution.

Recent methods increasingly combine multimodal large language models (MLLMs)~\cite{SAM4MLLM,SegAgent,SAM-Veteran} with foundation segmentation models such as SAM~\cite{SAM}. In these methods, MLLMs generate visual localization as spatial prompts for mask generation, exploiting the high-level reasoning and semantic understanding capabilities of MLLMs.

However, these methods still face two major limitations. First, when MLLMs generate spatial prompts~\cite{SAM4MLLM,SegAgent,SAM-Veteran} for the referred target in an autoregressive manner, coordinates are serialized into discrete text tokens. Such a representation fails to preserve the continuity of coordinates and their underlying spatial relationships, instead treating them as independent symbols, thereby increasing the risk of coarse or biased localization. Moreover, in complex scenes with densely distributed objects or highly similar appearances, one-pass methods that rely on MLLMs to localize the referent directly are prone to confusion with distractors. Such localization bias and alignment errors directly propagate incorrect spatial prompts to the downstream segmentation task, thereby degrading segmentation performance.

\begin{figure}[t]
    \centering
    \setlength{\abovecaptionskip}{2pt}
    \includegraphics[width=\columnwidth]{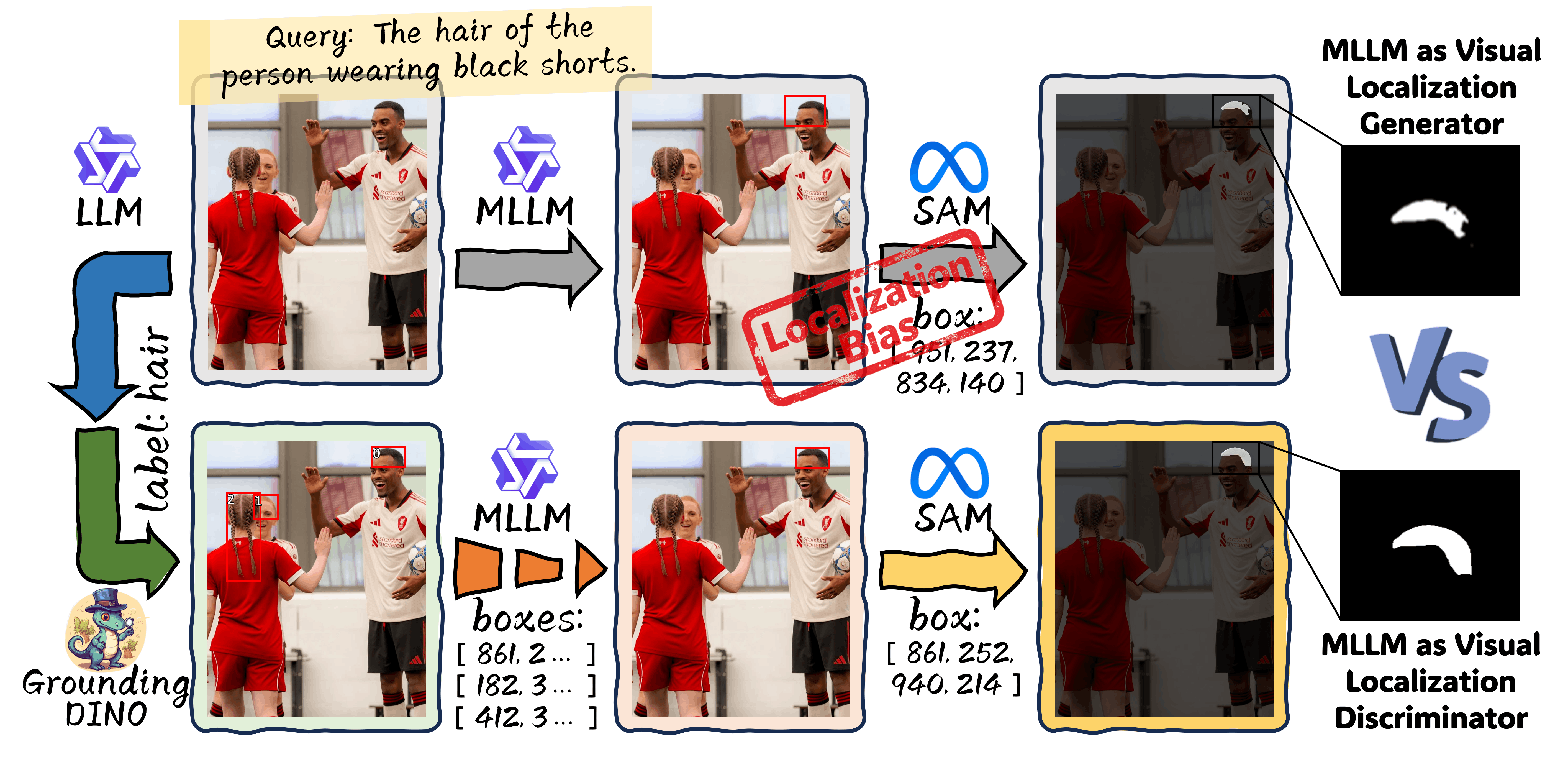}
    \caption{Generative visual localization by MLLMs is prone to localization errors and may propagate erroneous prompts to the downstream segmentation module. In contrast, discriminative reasoning mitigates this issue by selecting among detector-generated candidates.}
    \label{fig:intro}
    \vspace{-1.0em}
\end{figure}

Second, some methods introduce reasoning chains, such as Chain-of-Thought (CoT)~\cite{CoT} and multi-step iterative trajectories~\cite{SegAgent}, and use them to supervise the training of MLLMs, aiming to enhance their interpretability and reasoning capability in visual localization. However, these methods rely on high-quality reasoning-chain data, while valid reasoning paths for visual-semantic alignment are not unique~\cite{Self-consistency1}. Existing approaches often lack an effective mechanism for constructing and validating reasoning chains used for model training. Current methods typically obtain reasoning chains either through manual annotation or generation by powerful external MLLMs. The former is costly and difficult to scale, whereas the latter tends to produce logically inconsistent reasoning, which may cause the model to learn unstable or erroneous reasoning patterns.

Regarding these issues, we argue that a key limitation of many recent MLLM-based RES methods lies in treating visual localization as a generative coordinate prediction problem. RES can be better addressed by introducing a discriminative target selector, where the MLLM selects the referred object from a candidate space through screening and instance-wise verification, rather than generating localization coordinates in a one-pass manner. Moreover, reasoning-chain data used for training should not be accepted solely based on its surface plausibility. Instead, closed-loop consistency should be introduced as a reliability signal.

From this perspective, we propose DRAgent, an MLLM-driven discriminative target selection framework for RES. Rather than asking the MLLM to generate localization coordinates directly, DRAgent uses it as a visual-semantic discriminator over a detector-generated candidate space. An open-vocabulary detector~\cite{GroundingDINO} first constructs a high-recall candidate set, which serves as the decision space for the MLLM. The MLLM then performs two-stage discriminative reasoning (DR), consisting of candidate screening and instance-wise visual verification, to identify the candidate that best matches the referring expression. The selected bounding box is finally fed into SAM~\cite{SAM} as a spatial prompt to generate the pixel-level mask. To further strengthen the MLLM's DR capability, we construct a self-consistency-filtered reasoning-chain data pipeline and use the resulting data for LoRA-based~\cite{LoRA} supervised fine-tuning (SFT).

In summary, our contributions are three-fold.

\begin{itemize}
    \item We propose DRAgent, a RES framework that reformulates visual localization from coordinate generation into candidate-space DR, where the MLLM serves as a target discriminator over detector-generated candidates.
    \item We introduce a two-stage DR mechanism and a self-consistency-filtered reasoning-chain data pipeline, improving MLLM-driven target discrimination from the inference and training perspectives, respectively.
    \item Extensive experiments demonstrate that DRAgent achieves competitive performance on the RefCOCO, RefCOCO+, and RefCOCOg benchmarks.
\end{itemize}

\section{Related Work} 
\subsection{MLLMs for RES} 

Early RES studies explored task-specific architectures for aligning referring expressions with target regions. LAVT~\cite{LAVT} injects linguistic features into the visual backbone through a language-aware vision transformer. ReLA~\cite{ReLA} strengthens relation-aware cross-modal modeling, while PolyFormer~\cite{PolyFormer} formulates referring image segmentation as sequential polygon generation. However, they still struggle with fine-grained expression-region alignment in complex scenes.

To address these limitations, recent studies incorporate MLLMs into end-to-end RES frameworks to enhance open-vocabulary semantic understanding and cross-modal modeling capabilities. For example, LISA~\cite{LISA} connects an MLLM with SAM through the special \texttt{<SEG>} token, whose hidden-state embedding serves as a prompt to guide mask generation. GSVA~\cite{GSVA} and OpenWorldSAM~\cite{OpenWorldSAM} employ an MLLM as a multimodal feature encoder to provide semantically fused representations for the mask decoder. These methods have substantially improved cross-modal alignment. However, most of them primarily treat MLLMs as feature encoders~\cite{GLaMM,PixelLM} and do not fully exploit their strong reasoning capabilities. 

\subsection{MLLMs as Visual Localization Generators in RES} 

Some recent works have begun to use MLLMs as visual localization generators, producing bounding boxes, point prompts, or combinations thereof to guide external segmentation models for target mask generation. For instance, SAM-Veteran~\cite{SAM-Veteran} uses multi-round reasoning with an MLLM to generate and update point- and box-based coordinate prompts, which drive external segmentation tools to progressively refine the mask. In SegAgent~\cite{SegAgent}, the MLLM reads the current mask state together with the textual description and iteratively generates positive and negative point coordinates to simulate human collaboration with interactive segmentation tools, thereby refining the segmentation result step by step. Although such methods better exploit the explicit reasoning ability of MLLMs, generative visual localization for the referent~\cite{SAM4MLLM,DPAD} remains prone to localization bias and alignment errors, resulting in unreliable spatial prompts. 

In contrast to visual localization generator methods, DRAgent reformulates the role of the MLLM as a target discriminator. Candidate proposals are first produced by an open-vocabulary detector, while the MLLM is responsible only for discriminative selection over the candidate space. Specifically, DRAgent introduces a two-stage DR mechanism that combines candidate screening with instance-wise visual verification, enabling the MLLM to judge visual-semantic consistency rather than generate coordinates directly. This design decouples candidate generation from target discrimination and shifts the core MLLM function from spatial prompt prediction to reliable target verification.

\section{Method}

\begin{figure*}[t]
    \centering
    \setlength{\abovecaptionskip}{2pt}
    \includegraphics[width=\textwidth]{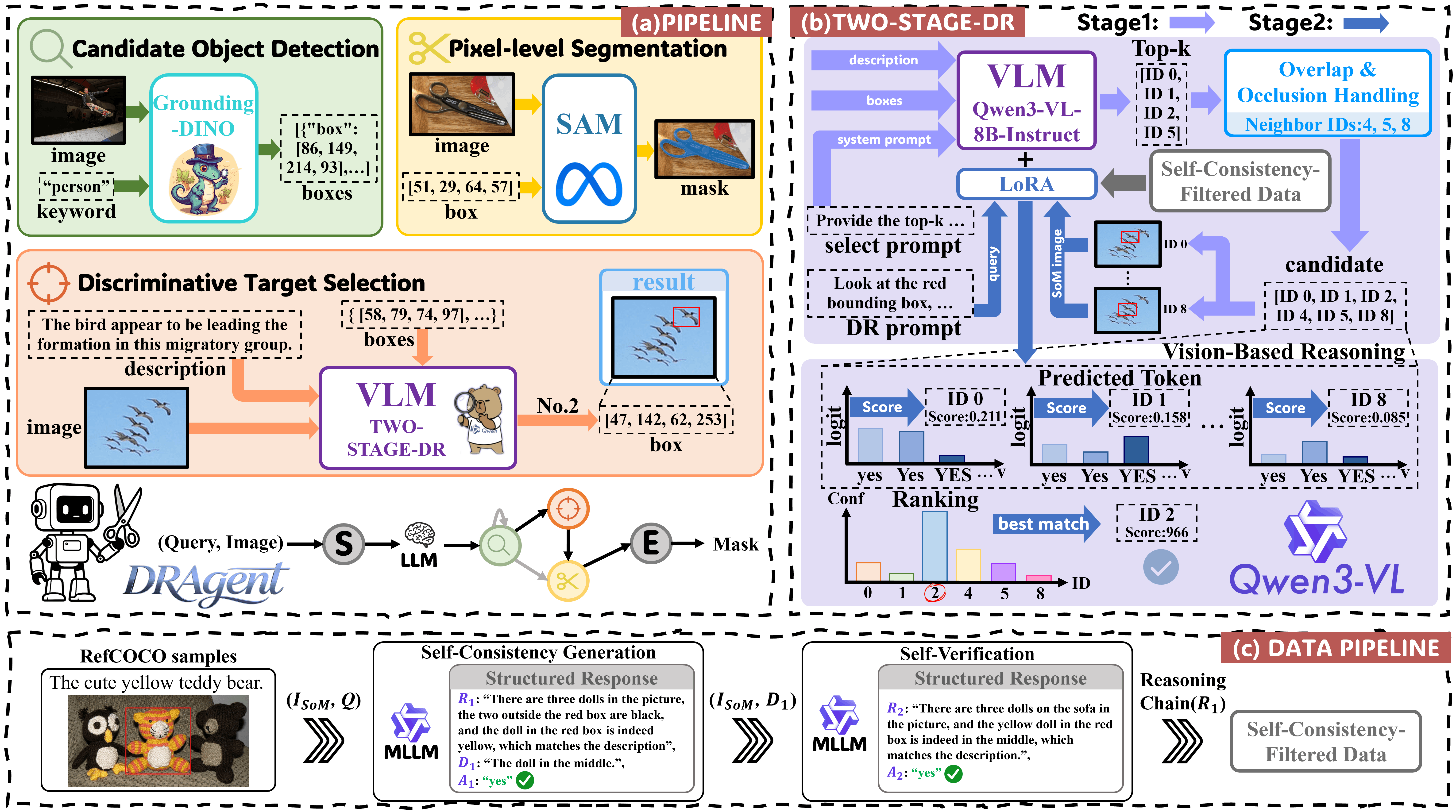}
    \caption{(a) DRAgent divides RES into three stages: candidate object detection, discriminative target selection, and pixel-level segmentation. (b) Two-stage DR mechanism for the discriminative target selection module. (c) Reasoning-chain data pipeline based on self-consistency filtering for SFT.}
    \label{fig:framework}
    \vspace{-1.0em}
\end{figure*}

DRAgent decomposes the RES task into three stages: candidate object detection, discriminative target selection, and pixel-level segmentation. Within this pipeline, the MLLM performs robust target selection through a two-stage DR mechanism, and its discriminative capability is further enhanced through fine-tuning with self-consistency-filtered reasoning-chain data.

\subsection{Overview of DRAgent}
\label{sec:overview}
We formulate the task pipeline of DRAgent as a multi-stage composite function. Given an input image $\mathcal{I}$ and a text query $\mathcal{Q}$, the goal is to generate a pixel-level mask $\mathcal{M}$.

\noindent\textbf{Candidate Object Detection.}
This stage constructs a high-recall discrete candidate space that serves as the decision space for subsequent discriminative target selection. Specifically, we first use an LLM to extract the target category from the text query $\mathcal{Q}$, denoted as $\mathcal{L}=\mathrm{LLM}(\mathcal{Q})$. We then feed it into an open-vocabulary detector, instantiated as Grounding DINO~\cite{GroundingDINO} in this work, to obtain a candidate set $\mathcal{O}=\{o_1,o_2,\ldots,o_n\}$, where each candidate $o_i$ is represented by a bounding box $[x_1^i,y_1^i,x_2^i,y_2^i]$.

\noindent\textbf{Discriminative Target Selection.}
This module uses the MLLM as a target discriminator to select the candidate $o_t$ from $\mathcal{O}$ that best matches $\mathcal{Q}$. The selection is performed through the two-stage DR mechanism detailed in Section~\ref{sec:dr}, rather than through direct coordinate generation.

\noindent\textbf{Pixel-Level Segmentation.}
After selecting the target $o_t$ referred to by the text query $\mathcal{Q}$, a pixel-level segmentation model, instantiated as SAM~\cite{SAM} in this work, is applied to generate the corresponding mask $\mathcal{M}$.

\noindent\textbf{Overall Pipeline Formulation.}
Taken together, the entire task can be represented as the following composite mapping:
\begin{equation}
\mathcal{M}=f_{\mathrm{seg}}\bigl(\mathcal{I},\Phi_{\mathrm{select}}(\mathcal{I},\mathcal{Q},g_{\mathrm{det}}(\mathcal{I},\mathrm{LLM}(\mathcal{Q}))))\bigr),
\end{equation}
where $g_{\mathrm{det}}$ denotes the open-vocabulary detector, $\Phi_{\mathrm{select}}$ denotes the core discriminative target selection module, and $f_{\mathrm{seg}}$ maps the final decision back to the pixel space.

\subsection{Two-Stage Target Selection}
\label{sec:dr}

Although the detector provides a candidate space for the MLLM, directly asking the MLLM to select the target from all candidates in one pass is prone to confusion with distractors in complex scenes. To make the MLLM's decision process more stable, we formulate target selection as a cascaded two-stage DR mechanism: the MLLM first performs candidate screening to reduce the search space, and then conducts instance-wise verification to identify the referred target through finer visual-semantic comparison.

\noindent\textbf{Candidate-Level Preliminary Screening.}
To reduce the search space while preserving candidate recall, we first visualize detected candidates on $\mathcal{I}$ with indexed Set-of-Mark (SoM)~\cite{SoM} bounding boxes. Based on this candidate-marked image, the MLLM selects the top-$K$ candidates from $\mathcal{O}$ that best match $\mathcal{Q}$. Because dense object layouts can impair the MLLM's judgment, we further supplement these top-$K$ candidates with candidates whose Intersection over Union (IoU) with the selected boxes exceeds a threshold $\gamma$, yielding the second-stage candidate set $\mathcal{O}_{\mathrm{top}}$.

\noindent\textbf{Discriminative Verification via Verification Score.}
To further reduce ambiguity among retained candidates, the second stage performs instance-wise visual verification. Each candidate $o_i \in \mathcal{O}{\mathrm{top}}$ is rendered with a SoM bounding box, denoted as $\mathcal{I}{\mathrm{SoM}}^i$. We design a task-specific instruction prompt, $\mathrm{Prompt}_{\mathrm{DR}}(\mathcal{Q})$, to elicit a structured visual response:

\begin{quote}
\small
``Look at the red bounding box. Does it tightly enclose the object described as `\{$\mathcal{Q}$\}'? Answer yes or no.''
\end{quote}

We perform candidate verification through vision-based discriminative reasoning, leveraging the visual question answering capability of the MLLM to determine from $\mathcal{I}_{\mathrm{SoM}}^i$ whether the candidate corresponds to the queried referent. This alleviates the mismatch between textual coordinate decoding and visual-spatial grounding by constructing an alignment score directly from visual evidence.

Concretely, let $\mathcal{V}$ denote the output vocabulary of the MLLM. Given the multimodal input $\mathcal{X}_i=\{\mathcal{I}_{\mathrm{SoM}}^i,\mathrm{Prompt}_{\mathrm{DR}}(\mathcal{Q})\}$, conventional generative decoding makes a direct decision based on the output:

\begin{equation}
\hat{v}_i=\arg\max_{v\in\mathcal{V}}\mathbb{P}_{\theta}(y_1=v\mid\mathcal{X}_i)
\end{equation}
where $y_1$ denotes the first generated answer token. This strategy is easily affected by an ambiguous probability distribution over answer tokens. To obtain a more stable and discriminative verification signal, we define the verification score of each candidate based on the relative strength between the positive and negative answer token sets:
\begin{equation}
S_i =\frac{\exp(z_{i,\mathcal{V}_{\mathrm{pos}}})}{\exp(z_{i,\mathcal{V}_{\mathrm{pos}}})+\exp(z_{i,\mathcal{V}_{\mathrm{neg}}})}.
\end{equation}
where $\mathcal{V}_{\mathrm{pos}}$ and $\mathcal{V}_{\mathrm{neg}}$ contain the token IDs corresponding to positive answer variants (``Yes'', ``yes'', and ``YES'') and negative answer variants (``No'', ``no'', and ``NO''), respectively. The terms $z_{i,\mathcal{V}_{\mathrm{pos}}}$ and $z_{i,\mathcal{V}_{\mathrm{neg}}}$ denote the average logits over the corresponding token sets. For example,
\begin{equation}
z_{i,\mathcal{V}_{\mathrm{pos}}}=\frac{1}{\lvert \mathcal{V}_{\mathrm{pos}}\rvert}\sum_{v\in\mathcal{V}_{\mathrm{pos}}}\mathrm{Logit}_{\theta}(v\mid\mathcal{X}_i).
\end{equation}
The term $z_{i,\mathcal{V}_{\mathrm{neg}}}$ is defined analogously. The logits are extracted at the first decoding step under a constrained yes/no answer format.
The final target $o_t$ is selected as the candidate with the highest verification score:
\begin{equation}
t=\arg\max_{i:o_i\in\mathcal{O}_{\mathrm{top}}}S_i.
\end{equation}
Here, $S_i$ is used as a relative ranking score rather than as a calibrated probability. 
It provides a signal for comparing the visual-semantic matching of different candidates.

\subsection{Data Pipeline and Training Strategy}
\label{sec:data}
\noindent\textbf{Reasoning-Chain Data Pipeline.}
To obtain reliable reasoning-chain supervision for model training, we construct a self-consistency-filtered data pipeline that filters out reasoning chains that are merely linguistically plausible but not grounded in correct visual target discrimination.

\paragraph{Stage 1. Self-Consistency Generation}
We first render the target region, which may correspond to either the referred target or a hard negative distractor, with an SoM prompt to obtain $\mathcal{I}_{\mathrm{SoM}}$, and then feed $(\mathcal{I}_{\mathrm{SoM}}, \mathcal{Q})$ into an MLLM and constrain its output to a structured response format using a task instruction prompt $\mathrm{Prompt}_{\mathrm{CoT}}(\mathcal{Q})$:

\begin{quote}
\small
``Look at the image. Does it accurately match \{$\mathcal{Q}$\}? Output JSON strictly as \{\texttt{"}reason\texttt{"}: \texttt{"}...\texttt{"}, \texttt{"}description\texttt{"}: \texttt{"}...\texttt{"}, \texttt{"}answer\texttt{"}: \texttt{"}Yes/No\texttt{"}\}.''
\end{quote}

The MLLM outputs a structured response containing a reasoning chain $\mathcal{R}_1$ which represents the reasoning process with respect to the query $\mathcal{Q}$, a description $\mathcal{D}_1$ which describes the visual content of the boxed candidate region, and an answer $\mathcal{A}_1$, which represents the decision result.

\paragraph{Stage 2. Self-Verification}

To avoid reasoning chains that are self-consistent only at the linguistic level but inconsistent with the visual evidence, we design a closed-loop verification~\cite{Self-consistency2} mechanism based on description reconstruction. Specifically, we use the $\mathcal{D}_1$ generated in Stage~1 to construct a new verification query $\mathrm{Prompt}_{\mathrm{CoT}}(\mathcal{D}_1)$, which is then fed back into the MLLM for closed-loop verification. The model again produces a structured response containing the reasoning chain $\mathcal{R}_2$ and answer $\mathcal{A}_2$. A reasoning chain is retained only if the first-stage answer $\mathcal{A}_1$ matches the ground-truth answer, and the second-stage answer $\mathcal{A}_2$ confirms the reconstructed description with ``Yes''. Otherwise, the sample is discarded from the training set.

\begin{table*}[t]
\centering
\caption{Comparison with existing state-of-the-art methods on RES benchmarks. The evaluation metric is cIoU (\%). The best and second-best results are highlighted in bold and underlined, respectively.}
\label{tab:res}

\scriptsize
\setlength{\tabcolsep}{3.0pt}
\renewcommand{\arraystretch}{1.05}

\begin{tabularx}{\textwidth}{L{0.155\textwidth} C{0.095\textwidth} *{8}{Y}}
\toprule
\multirow{2}{*}{Method} 
& \multirow{2}{*}{Venue}
& \multicolumn{3}{c}{RefCOCO}
& \multicolumn{3}{c}{RefCOCO+}
& \multicolumn{2}{c}{RefCOCOg} \\
\cmidrule(lr){3-5}
\cmidrule(lr){6-8}
\cmidrule(lr){9-10}
& & val & testA & testB & val & testA & testB & val(U) & test(U) \\

\midrule
\rowcolor{gray!12}
\multicolumn{10}{l}{\textit{Non-MLLM-based}} \\
LAVT~\cite{LAVT}       & CVPR'22    & 72.7  & 75.8  & 68.8  & 62.1  & 68.4  & 55.1  & 61.2  & 62.1 \\
SEEM~\cite{SEEM}       & NeurIPS'23 & \NA   & \NA   & \NA   & \NA   & \NA   & \NA   & 65.7  & \NA \\
ReLA~\cite{ReLA}       & CVPR'23    & 73.8  & 76.5  & 70.2  & 66.0  & 71.0  & 57.7  & 65.0  & 66.0 \\
PolyFormer~\cite{PolyFormer}  & CVPR'23    & 75.96 & 78.29 & 73.25 & 69.33 & 74.56 & 61.87 & 69.20 & 70.19 \\
CoHD~\cite{CoHD}       & ICCV'25    & 78.11 & 80.39 & 75.20 & 72.03 & 76.37 & 65.45 & 70.83 & 72.11 \\

\midrule
\rowcolor{gray!12}
\multicolumn{10}{l}{\textit{Implicit MLLM-based}} \\
LISA~\cite{LISA}         & CVPR'24    & 74.9  & 79.1  & 72.3  & 65.1  & 70.8  & 58.1  & 67.9  & 70.6 \\
GSVA~\cite{GSVA}         & CVPR'24    & 79.2  & 81.7  & \second{77.1} & 70.3  & 73.8  & 63.6  & \second{75.7} & \second{77.0} \\
GLaMM~\cite{GLaMM}        & CVPR'24    & 79.5  & \second{83.2} & 76.9  & 72.6  & 78.7  & 64.6  & 74.2  & 74.9 \\
PixelLM~\cite{PixelLM}      & CVPR'24    & 73.0  & 76.5  & 68.2  & 66.3  & 71.7  & 58.3  & 69.3  & 70.5 \\
PSALM~\cite{PSALM}        & ECCV'24    & \best{83.6} & \best{84.7} & \best{81.6} & 72.9 & 75.5 & \best{70.1} & 73.8 & 74.4 \\
OpenWorldSAM~\cite{OpenWorldSAM} & NeurIPS'25 & \NA   & \NA   & \NA   & \NA   & \NA   & \NA   & 74.0  & \NA \\
POPEN~\cite{POPEN}        & CVPR'25    & 79.3  & 82.0  & 74.1  & 73.1  & 77.0  & 65.1  & 75.4  & 75.6 \\
Dr.Seg~\cite{Dr.Seg}        & CVPR'26    & \NA   & 80.2  & \NA   & \NA   & 76.8  & \NA   & \NA   & 74.2 \\

\midrule
\rowcolor{gray!12}
\multicolumn{10}{l}{\textit{MLLMs as Explicit Visual Localization Generators}} \\
SAM4MLLM~\cite{SAM4MLLM}    & ECCV'24    & \second{79.8} & 82.7 & 74.7 & \best{74.6} & \best{80.0} & \second{67.2} & 75.5 & 76.4 \\
SAM-R1~\cite{SAM-R1}      & NeurIPS'25 & \NA   & 79.2  & \NA   & \NA   & 74.7  & \NA   & \NA   & 73.1 \\
SegAgent~\cite{SegAgent}    & CVPR'25    & 79.69 & 81.35 & 76.57 & 72.49 & 75.80 & 66.89 & 75.11 & 75.20 \\
DPAD~\cite{DPAD}        & CVPR'26    & \NA   & 79.3  & \NA   & \NA   & 74.7  & \NA   & \NA   & 72.6 \\
SAM-Veteran~\cite{SAM-Veteran} & ICLR'26    & \NA   & 80.8  & \NA   & \NA   & 76.6  & \NA   & \NA   & 73.4 \\

\midrule
\rowcolor{blue!6}
DRAgent & this work 
& 77.33 & 80.73 & 76.34 
& \second{73.14} & \second{78.79} & 64.29 
& \best{76.78} & \best{77.54} \\

\bottomrule
\end{tabularx}
\vspace{-1.0em}
\end{table*}

\noindent\textbf{Training Strategy.}
To enhance the MLLM's capability for visual-semantic discrimination while keeping the training cost manageable, we perform LoRA-based SFT~\cite{LoRA} using the filtered reasoning-chain data.

\section{Experiments}

\subsection{Experimental Settings}
\noindent\textbf{Implementation Details.}
We implement DRAgent using Grounding DINO~\cite{GroundingDINO} as the candidate object detector, Qwen3-VL-8B-Instruct~\cite{Qwen} as the MLLM-based target discriminator, and SAM~\cite{SAM} as the pixel-level segmentation model. Qwen3-VL-8B-Instruct is loaded with 4-bit quantization for efficient fine-tuning. LoRA~\cite{LoRA} is applied to both the visual and language layers, with both the rank and $\alpha$ set to 16. The maximum sequence length is set to 2048, the learning rate to $5\times10^{-6}$, and the weight decay to 0.01. Each training sample consists of a candidate-marked image and a text prompt, and the model outputs a short reasoning process together with the final verification answer.

\noindent\textbf{Benchmarks and Datasets.}
We conduct experiments on three standard RES benchmarks: RefCOCO, RefCOCO+, and RefCOCOg~\cite{RefCOCO/+,RefCOCOg}. For training, we use only 6K randomly sampled instances from the RefCOCO training set to construct the reasoning-chain data.

\noindent\textbf{Evaluation Metrics.}
Following prior work on text-guided segmentation~\cite{LAVT,LISA,GSVA,PSALM}, we adopt cumulative Intersection over Union (cIoU) as the evaluation metric, defined as the ratio between the accumulated intersection and the accumulated union over the entire dataset.

\noindent\textbf{Baselines.}
We compare DRAgent against three representative categories of methods on the RES benchmarks to provide a representative basis for comparison:
\begin{enumerate}
    \item Non-MLLM-based, which align visual and textual features through convolutional networks, Transformers, or related architectures, and directly predict the mask.
    \item Implicit MLLM-based, which use MLLMs for cross-modal fusion and perform mask prediction in an end-to-end manner.
    \item MLLMs as explicit visual localization generators, which use MLLMs to generate spatial prompts for foundation segmentation models to produce masks.
\end{enumerate}

\begin{table*}[t]
\centering
\scriptsize
\setlength{\tabcolsep}{2.8pt}
\renewcommand{\arraystretch}{1.05}

\begin{minipage}[t]{0.32\textwidth}
\vspace{0pt}
\centering
\caption{Ablation study on core components.}
\label{tab:ablation}
\vspace{0.4em}
\begin{tabularx}{\linewidth}{@{}Xccc@{}}
\toprule
\multirow{2}{*}{Setting} & \multicolumn{2}{c}{RefCOCOg} & \multirow{2}{*}{Avg.} \\
\cmidrule(lr){2-3}
& val(U) & test(U) & \\
\midrule
Baseline & 66.50 & 67.17 & 66.84 \\
+2Stage-DR & 74.13 & 76.70 & 75.42 \\
+2Stage-DR+SFT w/o CoT & 72.35 & 72.66 & 72.51 \\
+2Stage-DR+SFT w/ Raw CoT & 74.26 & 74.91 & 74.59 \\
DRAgent (Full) & 76.78 & 77.54 & 77.16 \\
\bottomrule
\end{tabularx}
\end{minipage}
\hfill
\begin{minipage}[t]{0.32\textwidth}
\vspace{0pt}
\centering
\caption{Parameter analysis of $K$.}
\label{tab:k}
\vspace{0.4em}
\begin{tabular*}{\linewidth}{@{\hspace{0.8em}}c@{\extracolsep{\fill}}ccc@{\hspace{0.8em}}}
\toprule
\multirow{2}{*}{$K$} & \multicolumn{2}{c}{RefCOCOg} & \multirow{2}{*}{Avg.} \\
\cmidrule(lr){2-3}
& val(U) & test(U) & \\
\midrule
2 & 73.76 & 75.95 & 74.86 \\
3 & 74.54 & 77.83 & 76.19 \\
4 & 76.78 & 77.54 & 77.16 \\
5 & 76.69 & 77.32 & 77.01 \\
6 & 75.87 & 77.24 & 76.56 \\
\bottomrule
\end{tabular*}
\end{minipage}
\hfill
\begin{minipage}[t]{0.32\textwidth}
\vspace{0pt}
\centering
\caption{Parameter analysis of $\smash{\gamma}$.}
\label{tab:gamma}
\vspace{0.4em}
\begin{tabular*}{\linewidth}{@{\hspace{0.8em}}c@{\extracolsep{\fill}}ccc@{\hspace{0.8em}}}
\toprule
\multirow{2}{*}{$\gamma$} & \multicolumn{2}{c}{RefCOCOg} & \multirow{2}{*}{Avg.} \\
\cmidrule(lr){2-3}
& val(U) & test(U) & \\
\midrule
40\% & 74.55 & 76.96 & 75.76 \\
50\% & 76.36 & 77.47 & 76.92 \\
60\% & 76.78 & 77.54 & 77.16 \\
70\% & 77.03 & 77.25 & 77.14 \\
80\% & 76.20 & 76.97 & 76.59 \\
\bottomrule
\end{tabular*}
\end{minipage}

\vspace{-1.0em}
\end{table*}

\subsection{Main Results}


As shown in Table~\ref{tab:res}, DRAgent achieves competitive overall results on RefCOCO, RefCOCO+, and RefCOCOg. Notably, it obtains the best performance on RefCOCOg, reaching 76.78\% on val(U) and 77.54\% on test(U), outperforming both implicit MLLM-based methods and explicit visual localization generator approaches. On RefCOCO and RefCOCO+, DRAgent does not consistently surpass the strongest methods, but remains competitive with recent explicit localization-generator methods.

These results support the effectiveness of using the MLLM as a discriminative target selector rather than a coordinate generator. Compared with methods that rely on implicit multimodal fusion, DRAgent explicitly exposes candidate objects to the MLLM and requires target-level discrimination. Compared with visual localization generator methods, DRAgent avoids directly decoding coordinates or points as textual outputs and instead performs target selection within a detector-generated candidate space. This design is particularly beneficial for RefCOCOg, where longer and more compositional expressions require fine-grained visual-semantic comparison among candidate objects.

\subsection{Ablation Study}

We conduct a systematic ablation study on the RefCOCOg dataset, with the detailed settings and results reported in Table~\ref{tab:ablation}. The baseline uses Grounding DINO~\cite{GroundingDINO} to generate candidates, a pretrained MLLM to select the target in a single stage, and SAM~\cite{SAM} to generate the final mask. We progressively incorporate the two-stage DR mechanism, SFT under different supervision settings, and filtered reasoning-chain data, in order to evaluate the contributions of each component as well as their joint effect.

The baseline achieves an average cIoU of 66.84\% on RefCOCOg, indicating that a single-stage pretrained MLLM selector is not sufficient to reliably identify the referred target from detector-generated candidates. After introducing the two-stage DR mechanism, the average cIoU increases to 75.42\%. This improvement can be attributed to the design of narrowing the candidate space through screening and then conducting instance-wise verification, which reduces the interference from irrelevant candidates while allowing the MLLM to focus on fine-grained visual-semantic matching. The fine-tuning variants show the influence of supervision design. Applying SFT without CoT supervision obtains an average cIoU of 72.51\%, which is lower than the two-stage DR variant. This suggests that final-answer supervision alone provides limited information about how the MLLM should compare candidates and make discriminative decisions. With unfiltered raw CoT data, the average cIoU reaches 74.59\%, indicating that reasoning-chain supervision can provide intermediate guidance for learning target discrimination. However, raw CoT data may still contain visually inconsistent or weakly grounded reasoning, which limits its effectiveness. The full DRAgent achieves the best average cIoU of 77.16\% with self-consistency-filtered reasoning-chain data, suggesting that consistency-based filtering provides more reliable supervision by retaining reasoning chains whose decisions are aligned with visual evidence. Overall, the two-stage DR mechanism improves the inference-time selection process, while self-consistency-filtered reasoning-chain data improves the training signal for discriminative target selection.

\begin{figure}[t]
    \centering
    \setlength{\abovecaptionskip}{2pt}
    \includegraphics[width=\columnwidth]{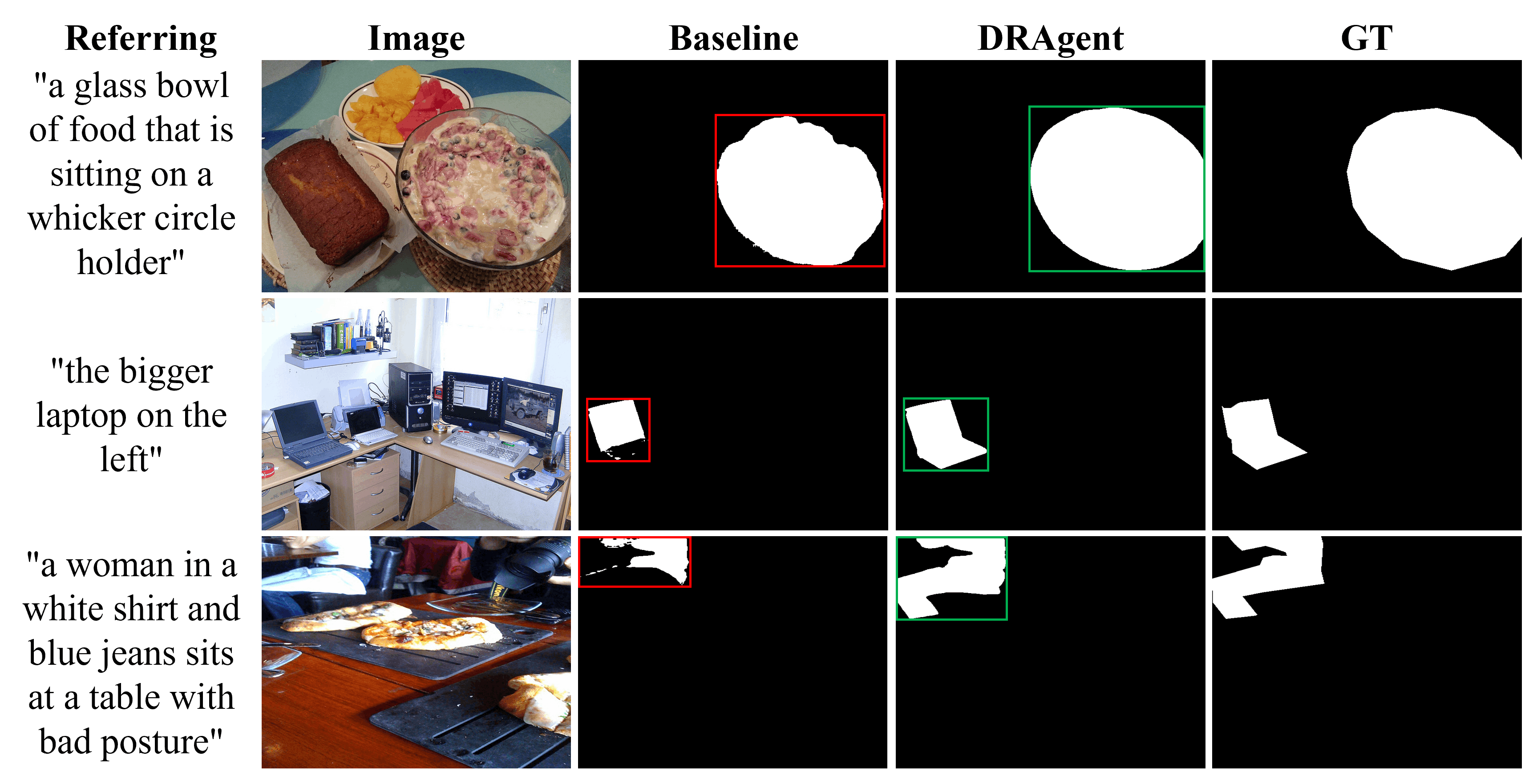}
    \caption{Visualization of qualitative results. Each row shows one RES example, with referring text, input image, baseline result, our DRAgent result, and ground truth (GT). Examples cover diverse challenging scenarios, and DRAgent achieves more accurate segmentation consistent with GT.}
    \label{fig:Visualization}
    \vspace{-1.0em}
\end{figure}

\subsection{Analysis}

During discriminative target selection, the number of initially selected candidates $K$ controls the size of the candidate subset. As shown in Table~\ref{tab:k}, the average cIoU increases as $K$ grows from 2 to 4, suggesting that retaining more candidates improves target recall and provides a richer decision space for subsequent verification. When $K$ is further increased, the average cIoU declines slightly, suggesting that an excessively large candidate set introduces more semantically irrelevant or visually confusable candidates, thereby increasing the difficulty of subsequent discriminative verification. Overall, $K=4$ achieves the best performance.

We also evaluate the overlap compensation threshold $\gamma$, which determines whether candidates overlapping with the initially selected boxes are retained. As shown in Table~\ref{tab:gamma}, increasing $\gamma$ from 40\% to 60\% improves the average cIoU, indicating that moderate overlap compensation helps preserve relevant neighboring candidates. However, further increasing $\gamma$ reduces performance, as an overly strict overlap threshold reduces the number of retained relevant candidates. Overall, $\gamma=60\%$ provides the best trade-off between candidate recall and noise suppression.

\section{Conclusion}

We present DRAgent, a MLLM-driven RES framework. DRAgent reformulates the role of the MLLM as a discriminative target selector over detected candidates. Through the two-stage DR mechanism, the MLLM first performs candidate screening and then conducts instance-wise verification to identify the referred target, which is subsequently converted into a mask by the segmentation model. We further introduce a self-consistency-filtered reasoning-chain data pipeline for LoRA-based fine-tuning, enhancing the MLLM's DR capability with more reliable supervision. Experiments on RefCOCO, RefCOCO+, and RefCOCOg demonstrate the effectiveness of the proposed framework.

\end{document}